\documentclass[journal]{IEEEtran}
\usepackage[pdftex]{graphicx}
\usepackage{cite}
\usepackage{color}
\usepackage[switch]{lineno}
\usepackage{graphicx}
\usepackage{subfigure}
\usepackage{url}
\usepackage{epstopdf}
\usepackage[table,xcdraw]{xcolor}

\usepackage{lineno}
\usepackage[justification=centering]{caption}
\usepackage{amsmath}
\usepackage{caption}
\usepackage{amsfonts}
\usepackage{booktabs}
\usepackage{multirow}
\usepackage{verbatim}
\usepackage{authblk}
\usepackage{bm}
\ifCLASSINFOpdf
\else
\fi
\begin{document}
%
\title{Cross-Modal Attentional Context Learning for RGB-D Object Detection}
%
%

\author{
        Guanbin~Li,
        Yukang~Gan,
        Hejun~Wu,       
        Nong~Xiao                
        and~Liang~Lin$^*$
\thanks{The first two authors contributed equally to this paper. The corresponding author is Liang Lin. G. Li, Y. Gan, H. Wu, N. Xiao and L. Lin are with the School of Data and Computer Science, Sun Yat-sen University, Guangzhou 510006, China.}
}

\maketitle

\begin{abstract}
Recognizing objects from simultaneously sensed photometric (RGB) and depth channels is a fundamental yet practical problem in many machine vision applications such as robot grasping and autonomous driving. In this paper, we address this problem by developing a Cross-Modal Attentional Context (CMAC) learning framework, which enables the full exploitation of the context information from both RGB and depth data. Compared to existing RGB-D object detection frameworks, our approach has several appealing properties. First, it consists of an attention-based global context model for exploiting adaptive contextual information and incorporating this information into a region-based CNN~(e.g., Fast RCNN) framework to achieve improved object detection performance. Second, our CMAC framework further contains a fine-grained object part attention module to harness multiple discriminative object parts inside each possible object region for superior local feature representation. While greatly improving the accuracy of RGB-D object detection, the effective cross-modal information fusion as well as attentional context modeling in our proposed model  provide an interpretable visualization scheme. Experimental results demonstrate that the proposed method significantly improves upon the state of the art on all public benchmarks. 
\end{abstract}


%
\IEEEpeerreviewmaketitle

\begin{figure*}
  \centering
  \includegraphics[width=0.9\textwidth]{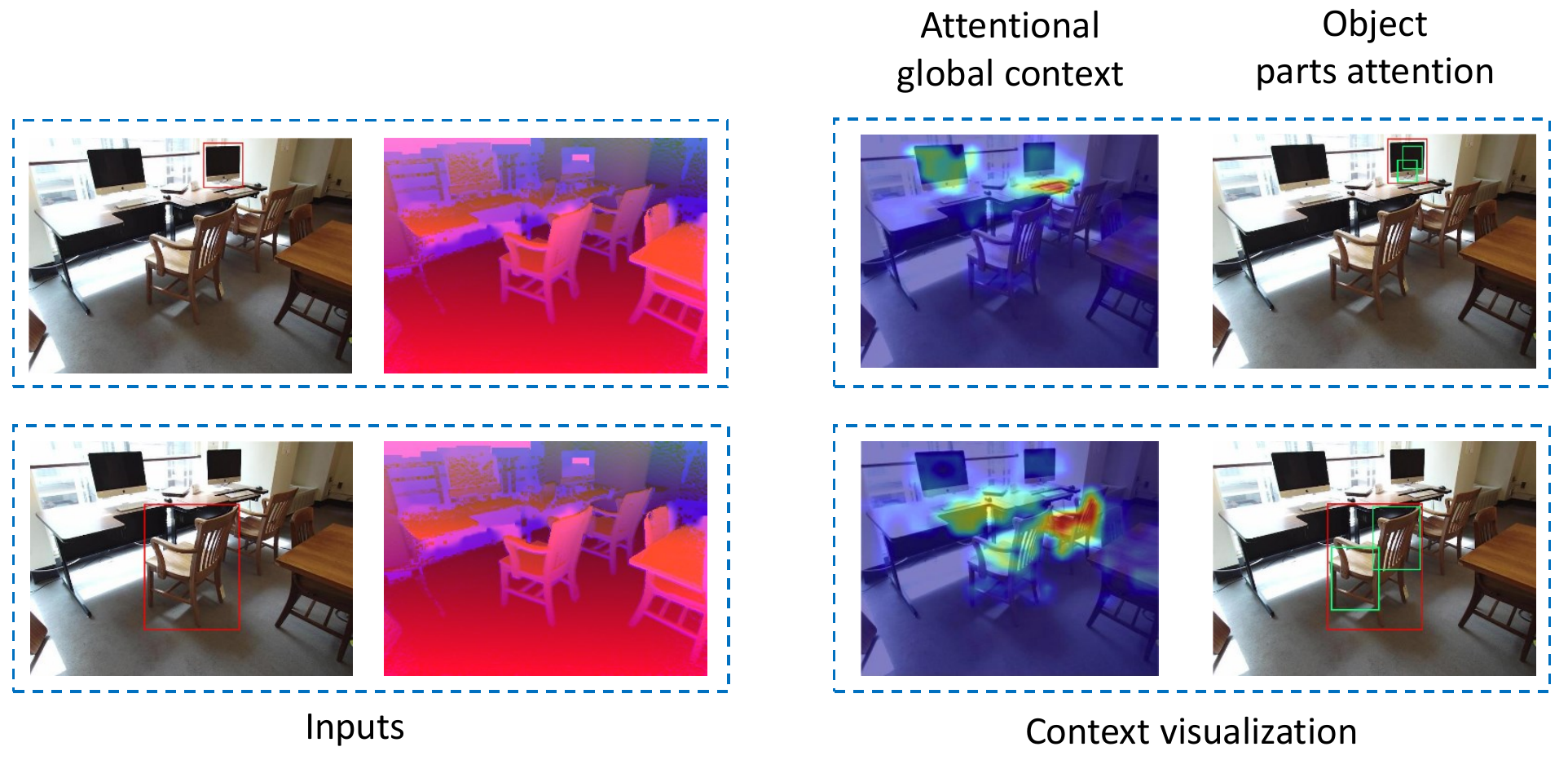}
  \caption {\label{figure:1} Example visualization results for global context and object part attention generated by our proposed CMAC model. For global context, information from relevant regions (the highlighted regions) of the object proposals is obtained through a recurrent attentional model. For local context, multiple parallel spatial transformers are utilized to exploit information from the discriminative parts (green rectangles) of the object proposals. Red rectangles indicate the object proposals.}
\end{figure*}

\section{Introduction}
%
%
%
%
\IEEEPARstart{R}GB-D object detection attempts to localize and classify objects within an image with depth information. It is one of the core technologies in the field of robotics application and can be beneficial to many intelligent tasks, including pose estimation \cite{hinterstoisser2012model, wang2016human}, content-based image retrieval \cite{wu2014hierarchical} and robot task planning \cite{schuster2015generating}. In recent years, the successful application of deep convolutional neural networks has pushed this research into a new phase and achieved very good results.

Most CNN-based RGB-D object detection frameworks are extended from RCNN-based object detectors \cite{girshick2014rich, fast-rcnn, ren2015faster} for RGB images. R-CNN-Depth \cite{gupta2014learning} is the first deep learning framework for RGB-D object detection that extends the R-CNN system~\cite{girshick2014rich} to take advantage of depth information by incorporating two parallel network streams for both RGB and depth modalities. 
 This two-stream pipeline later became the basis for many visual perception tasks in RGB-D images \cite{eitel2015multimodal, gupta2016cross, schwarz2015rgb, li2016lstm}. In this framework, the features from the RGB and depth modalities are computed independently and concatenated after applying fully connected layers for final proposal classification. However, this pipeline has its own limitations: (1)~Independent feature computation and simple feature concatenation ignore the correlation between the two modalities. (2)~Only information inside the object proposal is used for object classification, which neglects the auxiliary role of context information outside the bounding box in object classification.

In this paper, we propose a Cross Modal Attentional Context (CMAC) learning framework for RGB-D object detection that incorporates the consistency and complementary information between two diverse modalities~(RGB and depth), as well as an attentional model for global context mining and discriminative object part discovery. To exploit the correlation between RGB and depth modalities, the CMAC model employs a cross-modal feature fusion component to fuse the features extracted from the output feature maps of the two fully convolutional networks~(with different input sources). Instead of directly applying fused features to classification and object location refinement, our proposed CMAC model further learns attentional context and explores discriminative object parts based on the fused features. We believe that both the attentional global context and the discriminative parts attended inside each possible object region~(object proposal) are crucial for accurate RGB-D object detection. 

To capture the global context, our model employs a recurrent attention model that consists of multiple stacked Long Short-Term Memory (LSTM) units. The recurrent neural network is optimized to infer relevant regions for each given region proposal. As shown in Figure~\ref{figure:1}, the regions that are considered helpful for classification of the object proposal are highlighted. As can be seen, our proposed CMAC model can identify an adaptive global context for different object proposals~(\textit{i.e.,} the regions of the keyboard, parts of the table around the target monitor as well as the other monitor are highlighted when the input region proposal contains a monitor. When the input region proposal contains a chair, the regions including parts of the table and other chairs are assigned higher weights in the final classification.). Moreover, inspired by the fact that 
humans tend to quickly capture distinguishable parts for more accurate object classification judgment when observing objects with occluded regions, we propose to further incorporate a fine-grained object part attention module in our network framework. Considering the flexible attention mechanism and the excellent spatial manipulation ability of Spatial Transform Networks (STNs), we adopt multiple STNs in parallel to examine the discriminative parts located inside a specific object proposal for capturing local context. As illustrated in Figure~\ref{figure:1}, the CMAC model is able to successfully locate the most discriminative location that can differentiate an object's category~(\textit{i.e.,} the main screen and the base of the monitors, as well as the back and legs of the chairs). Acquiring such fine-grained object parts provides enhanced feature representations for region proposals.

In summary, the main contributions of the proposed CMAC model can be listed as follows:




\begin{itemize}
\item[-] We propose a novel Cross Modal Attentional Context~(CMAC) deep learning framework that effectively incorporates the correlated information between different modalities and successfully identifies useful contextual information both locally and globally for RGB-D object detection.
\item[-] An attention-based global context module, based on an LSTM network, is utilized to recurrently generate contextual information from a global view for each object proposal.
\item[-] Multiple spatial transform networks are adopted in parallel to localize discriminative object parts for accurate object recognition. 
\item[-] Extensive experiments on the SUNRGBD and NYUv2 datasets well demonstrate the effectiveness of the proposed CMAC model, which outperforms the state-of-the-art method \cite{gupta2016cross} by 3.7\% and 3.2\%, respectively, in terms of mAP. 
\end{itemize}

\begin{figure*}
  \centering
  \includegraphics[width=0.9\textwidth]{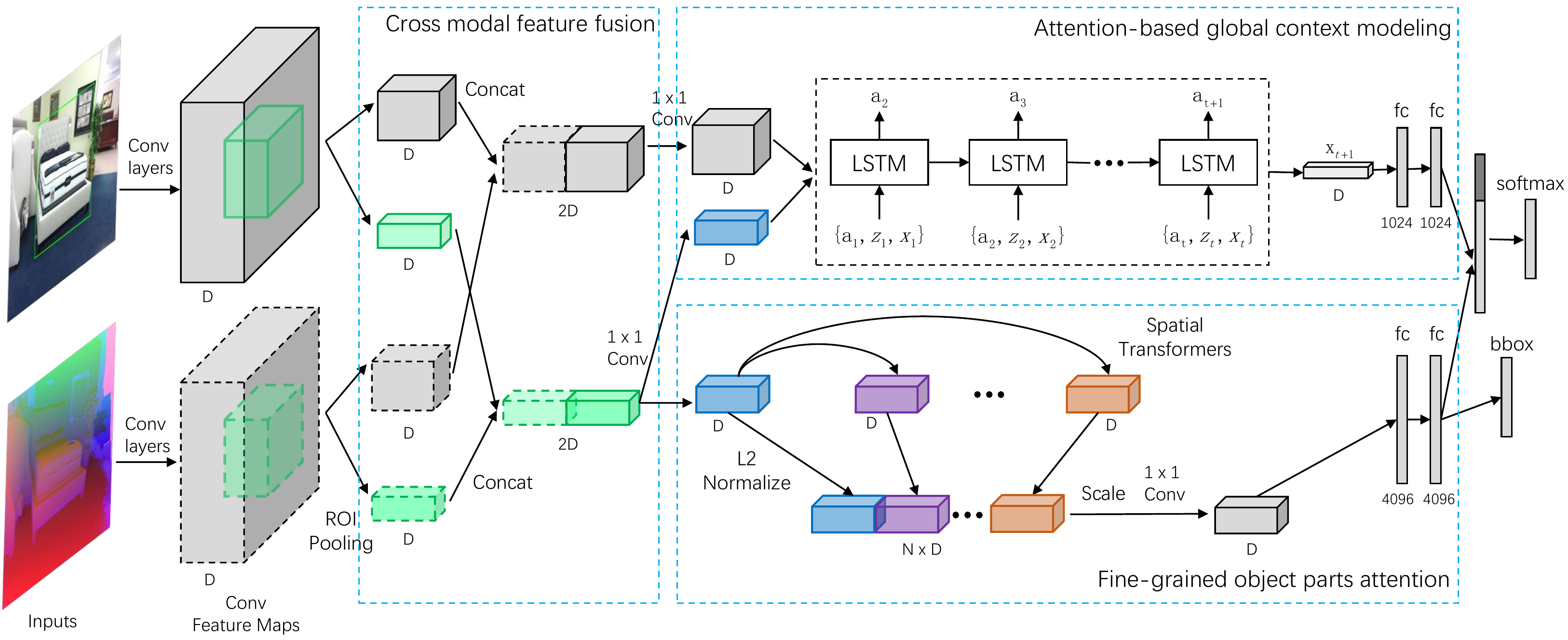}
  \caption {\label{fig:framework} The network architecture of our proposed cross-modal attentional context~(CMAC) learning framework. The input consists of one RGB image and one HHA image (geocentric encoding of the depth image). Our network framework is composed of four components: convolutional feature extraction, cross-modal feature fusion, attention-based global context modeling and fine-grained object part attention.}
\end{figure*}

\section{Related Work}
\subsection{Object Detection in RGB-D Images}
Object detection in RGB-D images has attracted increased attention because of the rapid development of affordable depth sensors and their diverse application scenarios. Many successful algorithms have been proposed to effectively exploit information from RGB-D data. \cite{bo2011depth} and \cite{lai2011large} took advantage of hand-designed features such as SIFT and multiple shape features in the depth channel for RGB-D object recognition. Schwarz \textit{et al}. \cite{schwarz2015rgb} utilized two-stream CNNs pre-trained on ImageNet to extract features from RGB-D images. While most work mainly focuses on the RGB modality, some recent work has been dedicated to improving the object detection performance by taking depth information into consideration. Gupta \textit{et al}.~\cite{gupta2014learning} proposed a geocentric embedding to convert each single-channel depth map into a three-channel depth image~(HHA image), in which they encoded each pixel with three channels of information, \textit{i.e.,} the height above the ground, the horizontal disparity and the angle with respect to gravity. They also introduced a generalized method for the R-CNN detector that can be applied to RGB-D images; they used large CNNs pre-trained on RGB images to extract features from HHA data. To learn rich representations for the depth modality, \cite{gupta2016cross} transferred supervisions from labeled RGB images to unlabeled depth images. In this paper, we follow \cite{gupta2014learning} and encode depth information into HHA images for improved feature learning and take the model in~\cite{gupta2016cross} as our compared baseline model.

Another core issue of RGB-D object detection is how to merge the features from different sources. Existing fusion strategies can be divided into two streams: (1) Early fusion \cite{blum2012learned, bo2011object, bo2011depth}, in which the depth channel is being treated as an extra channel to RGB images and is concatenated with the RGB channels for feature extraction. (2) Late fusion~\cite{eitel2015multimodal, gupta2014learning, spinello2012leveraging, gupta2016cross}, where features are separately learned for each modality and are concatenated at later stages for object classification. Our model is similar to the late fusion approach, but instead of directly concatenating features for classification, we apply the attention model to the fused features to learn a better global context and discriminative object parts to achieve more accurate object recognition.

\subsection{Context Information in Object Detection}
Context information has been applied in many methods to enhance the performance of object detection~\cite{carbonetto2004statistical,li2016visual, divvala2009empirical, heitz2008learning, hoiem2005geometric, li2018contrast,torralba2010using}. For instance, \cite{torralba2010using} exploited context from information about the entire scene  for  object detection and localization. \cite{heitz2008learning} explored contextual relationships between regions in an unsupervised manner, where objects are detected using a discriminative approach. Spatial support and geographic information are used as context clues in \cite{divvala2009empirical}. Context models have also been applied to deep-learning-based object detectors. \cite{li2017multi} proposed a group recursive learning approach to refine object proposals by incorporating semantic and spatial layout correlations of surrounding proposals. Chu et al.~\cite{chu2016deep} formulated a fully connected conditional random field (CRF) to incorporate the local appearance and the contextual information in terms of relationships among objects and the global scene based on contextual features generated by a convolutional neural network. Inside-Outside net (ION) \cite{bell2016inside} introduced spatial recurrent networks (RNNs) to integrate the contextual information outside the region of interest while utilizing skip pooling to extract fine-grained information from multiple low-level convolutional layers. Although our proposed model also explores global contextual information through recurrent networks, it explicitly learn to attend the most relevant regions of the object proposal by generating a weight map for each proposal. The weight map can well reveal the contextual region that corresponds to the final classification result. One the other hand, instead of directly extracting local features from the whole object bounding box, our model can achieve better object feature representations by recurrently discovering the most discriminative object parts inside the object proposal and performing part-level feature fusion.

\subsection{Recurrent Attention Models}
Recurrent attentional models have been widely incorporated in deep-learning-based computer vision tasks~\cite{bahdanau2014neural, li2016attentive, mnih2014recurrent, xu2015show} to achieve better performance. Bahdanau et al.~\cite{bahdanau2014neural} introduced recurrent attention to neural machine translation, which allows the model to adaptively attend to the most relevant part of a sentence.~\cite{mnih2014recurrent} adopted visual attention to dynamically select a sequence of regions and only processed the selected regions for efficient computation. A recent work in~\cite{xu2015show} used an LSTM-based attention model to learn a description of static images. More recently, an attention mechanism has also been applied to vision tasks for videos. For instance, \cite{yao2015describing} extended an attention model for video description and employed a temporal attention mechanism to model the dynamic temporal structure of videos. \cite{sharma2015action} optimized the attention model to attend to the relevant parts within a single frame and attached higher importance to them while performing action recognition.

The  work that is most relevant  to our proposed method is the attentive context proposed in~\cite{li2016attentive}, which also incorporated a recurrent attention model to exploit global contextual information. However, the attention model used in \cite{li2016attentive} generated a static attentive location map for all object proposals. Instead of utilizing a fixed attentive context, our model generates an attentional context feature adaptive to the input region proposals. Furthermore, we employ a fine-grained object part attention module to harness multiple discriminative object parts inside each object proposal for achieving a superior local feature representation. Experimental analysis in Sec.~\ref{sec:ablation} demonstrates that our method is more robust to background and inter-class noise.

\section{Framework}\label{sec:framework}
An overview of our framework is illustrated in Fig.~\ref{fig:framework}. Our RGB-D object detection system, which is based on cross modal attentional context learning, is composed of four components, including fully convolutional networks based feature extraction, cross-modal feature fusion, attention-based global context modeling and fine-grained object part attention. We term this network Cross-Modal Attentional Context~(CMAC) network. Specifically, given an RGB-D image, we first employ Multiscale Combinatorial Grouping~(MCG)~\cite{mcg} to generate a number of object proposals from RGB information and encode the original depth value to the three-channel HHA representation, as proposed in~\cite{gupta2014learning}. Following the benchmark object detection framework of Fast R-CNN~\cite{fast-rcnn}, our CMAC model takes as input an RGB image, an HHA image and corresponding object proposals to generate class labels as well as a refined bounding box for each object proposal. 

As shown in Fig.~\ref{fig:framework}, the feature extraction module is built on two separate fully convolutional sub-networks, including the VGG16 model~\cite{VGG} for RGB modality and AlexNet model~\cite{alexnet} for depth modality. The output of the last convolutional layer is being treated as our initial feature for object detection, therein including $D$ convolutional maps. The two fully convolutional sub-networks take as input the RGB image and the HHA image to generate the corresponding feature cube. Region-of-Interest (RoI) pooling operations are performed on the two feature cubes to obtain both global~(whole image) and local features~(object proposal) of the two modalities before being fed to a cross-modal feature fusion module. Moreover, both the fused global feature and the fused local feature are fed to a global context modeling module to obtain an attentional global context feature for the corresponding object proposal, while the fused local feature itself is also treated as an input for the fine-grained object part attention, which generates an embedded local feature. Finally, the concatenation of the global context feature and the embedded local feature are employed for final object detection, while local feature embedding is applied for further bounding box regression. 

\subsection{Cross-Modal Feature Fusion}
It has been widely verified that the RGB modality and depth modality are complementary, the combination of which can help to boost the RGB-D object detection performance~\cite{gupta2014learning,gupta2016cross}. In this paper, we exploit the features extracted from the two modalities for both global context modeling and local proposal feature description. Specifically, we design a simple yet effective sub-network to fuse features extracted from both modalities. For each object proposal, we extract a fixed-size feature representation using ROI pooling \cite{fast-rcnn} in both modalities, denoted as \textit{$\bm{F}_{l\_rgb}$} and \textit{$\bm{F}_{l\_depth}$}. We also apply a pooling operation to the output feature map of the last convolution layer of the two fully convolutional networks to generate fixed-size feature cubes, denoted as~\textit{$\bm{F}_{g\_rgb}$} and \textit{$\bm{F}_{g\_depth}$}, respectively. The feature fusion between RGB and depth modality can be represented by

\begin{equation}
  \textit{$\bm{F}_{l\_fused}$} = \textit{concat}(\textit{$\bm{F}_{l\_rgb}$},        \textit{$\bm{F}_{l\_depth}$})
\end{equation}

\begin{equation}
  \textit{$\bm{F}_{g\_fused}$} = \textit{concat}(\textit{$\bm{F}_{g\_rgb}$},        \textit{$\bm{F}_{g\_depth}$})
\end{equation} 

where \textit{$\bm{F}_{l\_fused}$} and \textit{$\bm{F}_{g\_fused}$} are the global context feature and local object proposal feature after fusion, respectively, and \textit{concat($\cdot$)} indicates the concatenation operation of feature representations along the channel axis.

In contrast to \cite{gupta2014learning, gupta2016cross, song2016deep}, which apply two independent CNNs to separately extract features from both modalities and directly perform simple concatenation for final classification, our cross-modal feature fusion operation is treated as a feature generation step for further global context modeling and local feature embedding before final classification. In the experiment section, we verify that our proposed cross-modal feature representation can help to produce more effective local and global context information, greatly improving the performance of the final classification.

\subsection{Attention-based Global Context Modeling}\label{sssec:global}
It is well known that contextual representation is crucial for accurate visual recognition~\cite{bell2016inside, li2015visual, li2016attentive, mottaghi2014role,li2017context,li2017instance}. Instead of directly obtaining fixed context information to assist in object detection~\cite{li2016attentive,mottaghi2014role}, we focus on exploiting adaptive context information for each object proposal. Specifically, we design a soft attention model  based on multi-layered RNNs with  LSTM units to spatially weight the features and generate an adaptive global context feature for each object proposal. Average pooling and max pooling operations over the feature map of the whole image can be considered as special cases of our method.

The attentional context model takes as input the concatenation of the global feature cube and that of the local feature cube before being fed to a $1 \times 1$ convolutional layer for feature embedding. The dimensions of the embedded global and local feature are denoted as $K\times K\times D$ (20$\times$20$\times 512$ in our experiments) and $S\times S\times D$ (7$\times$7$\times 512$ in our experiments), respectively. Based on these embedded feature cubes, the RNN model learns an attentional map of size \textit{K $\times$ K} to determine the effectiveness of the contextual region that may be beneficial to the object detection. 

Inspired by the LSTM-based soft attention model proposed in~\cite{sharma2015action}, we apply an LSTM network to generate a contextual attention map at every time step conditioned on the previous hidden state, the globally embedded feature vector as well as the local feature. Specifically, at each time-step $t$, we extract $K^2$ D-dimensional global feature vectors as well as $S^2$ local object proposal feature vectors. As in~\cite{sharma2015action}, we refer to these feature vectors as global feature slices and local feature slices, respectively, denoted as

\begin{table} \centering \renewcommand\arraystretch{1.2}
	\captionsetup{font={small}}
	\caption{\label{table:6} Detection results from different methods on SUNRGBD and NYUv2. AC-CNN* indicates our implementation of the RGB-D version of AC-CNN \cite{li2016attentive}. G and L  denote our proposed model incorporated with a single LSTM module~(G) or STN module~(L), respectively. (w/o fusion) and (w/ fusion) denote without and with multi-modal context fusion, respectively.}
	\begin{tabular}{c|cc|c|c}
	\toprule
	\multirow{2}{*}{\textbf{Method}} &
	\multirow{2}{*}{\textbf{G}} &
	\multirow{2}{*}{\textbf{L}} &
	\multicolumn{2}{c}{\textbf{mAP}} \\

	\cline{4-5}
		& & & \small{SUNRGBD} & \small{NYUv2} \\

	\midrule
	ST(baseline) \cite{gupta2016cross} &&& 43.8 & 49.1 \\
	AC-CNN* \cite{li2016attentive} &$\surd$ &$\surd$ & 45.4 & 50.2 \\
	
	\midrule
	
	\multirow{3}*{Ours (w/o fusion)} & &$\surd$ & 46.3 & 50.9 \\
	&$\surd$ & & 46.2 & 51.3 \\
	&$\surd$ &$\surd$ & 46.9 & 51.9 \\
	\midrule
	Ours (w/ fusion) &$\surd$ &$\surd$ & \textbf{47.5} & \textbf{52.3} \\
	
	\bottomrule
	\end{tabular}
\end{table}

\begin{table} \centering \renewcommand\arraystretch{1.2}
	\captionsetup{font={small}}
	\caption{\label{table:8} Comparison of exploiting global context using different methods on SUNRGBD and NYUv2}
	\begin{tabular}{c|c|c}
	\toprule
	\multirow{2}{*}{\textbf{Method}} &
	\multicolumn{2}{c}{\textbf{mAP}} \\

	\cline{2-3}
		& SUNRGBD & NYUv2 \\

	\midrule
	Average Pooling &44.3 & 49.4\\
	Fixed Attentive Context \cite{li2016attentive} & 44.8 & 49.7 \\
	Adaptive Attentive Context (Ours) & 46.2 & 51.3 \\
	
	\bottomrule
	\end{tabular}
\end{table}

\begin{equation}
\begin{cases}
G_t = \left [ G_{t,1},...,G_{t,K^2} \right ] & G_{t,i} \in \mathbb{R}^D\\ 
\cr
L_t = \left [ L_{t,1},...,L_{t,S^2} \right ] & L_{t,i} \in \mathbb{R}^D 
\end{cases}
\end{equation}

Each vertical column of $G_t$ and $L_t$ denotes the feature representation~(receptive field) in the input image. We follow the implementation of the LSTM network in~\cite{hochreiter1997long}, which is formulated as

\begin{equation}
\label{equ:3}
\left( 
  \begin{array}{lr}
  \bm{i}_{t} \\ 
  \bm{f}_{t} \\
  \bm{o}_{t} \\
  \bm{g}_{t}
  \end{array} 
\right) \\ 
= \\
\left( 
  \begin{array}{lr}
  \sigma \\ 
  \sigma \\
  \sigma \\
  tanh
  \end{array} 
\right) \\
\ \textit{\large{T}} \\
\ \left( 
  \begin{array}{lr}
  \bm{h}_{t-1} \\ 
  \bm{x}_{t} \\
  \bm{z}_{t} \\
  \end{array} 
\right),
\end{equation}
\begin{equation}
\label{equ:4}
  \bm{c}_{t} = \bm{f}_{t} \odot \bm{c}_{t-1} + \bm{i} \odot \bm{g}_{t}
\end{equation}
\begin{equation}
\label{equ:5}
  \bm{h}_{t} = \bm{o}_{t} \odot tanh(\bm{c}_{t})
\end{equation}

where $\bm{i}_{t}$, $\bm{f}_{t}$, $\bm{c}_{t}$, $\bm{o}_{t}$, and $\bm{h}_{t}$ are the input gate, forget gate, cell state, output gate and hidden state of the LSTM, respectively; $\bm{x}_{t}$ is the global context feature vector input to the LSTM at time step \textit{t}; the vector $\bm{z} \in \mathbb{R}^{D}$ is the local feature embedding of the object proposal with the global average pooling operation; $\textit{T} \in \mathbb{R}^{(2D + d) \times 4d}$ denotes a simple affine transformation with trainable parameters, where $d$ is the dimensionality of $i_t$, $f_t$, $c_t$ and $h_t$; and $\sigma$ and $\odot$  denote the logistic sigmoid activation and element-wise multiplication, respectively.

At each time step $t$, our LSTM model learns to predict a weight map $\alpha_{t+1}$ of size $K\times K$, where its value corresponds to the spatial attention that should be paid when performing proposal classification. The weight map \textit{$\alpha_{i}$} is computed by a multilayer perception $\phi$ conditioned on the previous hidden state \textit{$h_{t-1}$}. The spatial weight of $\alpha_{t}$ at location $i$ can thus be computed as follows: 

\begin{equation}
  e_{ti} = \phi(\bm{h}_{t-1}) 
\end{equation}

\begin{equation}
  {\alpha}_{ti} = \frac{exp(e_{ti})}{\sum_{k=1}^{K \times K}exp(e_{tk})}
\end{equation}
Based on the weight map, the global context feature vector $x$ at time step $t$ is computed as an average of the feature slices weighted according to $\alpha_{t}$, formulated as
\begin{equation}
  \bm{x}_{t} = \sum_{i=1}^{K^{2}}{{\alpha}_{t,i}\bm{F}_{g\_fused, i}}
\end{equation}
where $F_{g\_fused, i}$ is the $i^{th}$ global feature slice. Because the relevant regions are given higher weights, the global feature $\bm{x}_{t}$ will be dominated by features from these regions and hence provide more useful contextual information for more accurate object detection.

During the initialization stage, we follow the same strategy proposed in~\cite{show2015tell} for faster convergence. Specifically, we initialize the cell state $\bm{c}_t$ and the hidden state $\bm{h}_t$ of the LSTM network as 
\begin{equation}
  \bm{c}_0 = \textit{f}_{\textit{init, c}} \left(\frac{1}{K^2}\sum_{i=1}^{K^2}\bm{F}_{g\_fused, i} \right)
\end{equation}

\begin{equation}
  \bm{h}_0 = \textit{f}_{\textit{init, h}} \left(\frac{1}{K^2}\sum_{i=1}^{K^2}\bm{F}_{g\_fused, i} \right)
\end{equation}
where $\textit{f}_{\textit{init, c}}$ and $\textit{f}_{\textit{init, h}}$ are two multi-layer perceptions. The two initial values are applied to infer the initial weights $\alpha_1$ for the initialization of the global context feature $\bm{x}_1$.

As shown in Fig.~\ref{fig:framework}, the output of our LSTM model is a D-dimensional global context feature, which is further fed to two fully connected layers to produce the final feature representation, denoted as $F_G$.

\begin{figure}
  \centering
  \includegraphics[width=0.4\textwidth]{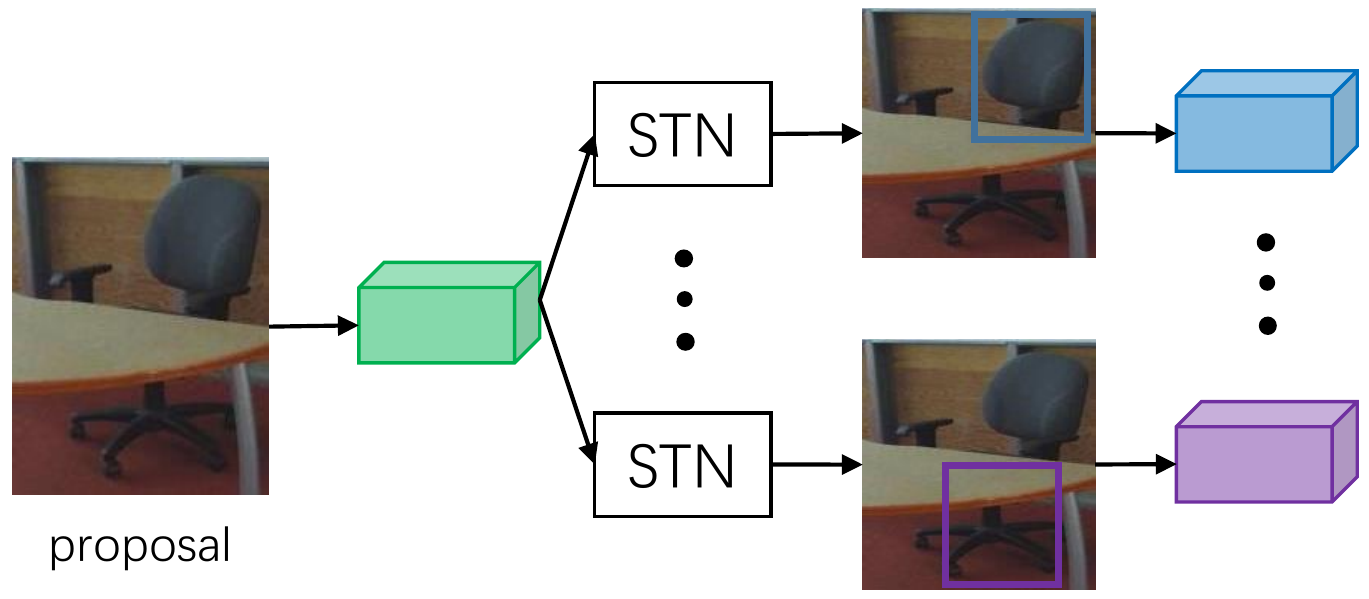}
  \caption {\label{fig:stn} Illustration of the STN module. The STN module takes the feature of the object proposal as input and attends to the most discriminative parts. The feature from these parts will subsequently serve as an enhanced local feature in object classification and bounding box regression.}
\end{figure}

\subsection{Fine-grained object part attention}
\label{sssec:local}
Because the local salient parts inside a specific object proposal play an important guiding role in judging the classification of an object (especially for partially occluded objects), we further propose to employ multiple STNs~\cite{jaderberg2015spatial} in parallel to infer discriminative object parts for each object proposal. The spatial transformer is a differential module that learns to spatially transform the input feature maps $U$ to the output feature maps $V$. A spatial transformer is applied in the following three steps. First, a localization network is employed to predict the affine transformation matrix $A_{\theta}$ to be applied to the input feature map. Second, $A_{\theta}$ is being applied to create a sampling grid in $U$ by the grid generator. Finally, a sampler is adopted to produce the output maps sampled from the regions of input maps at the sampling grid. As shown in Figure \ref{fig:stn}, we train each transformer to automatically attend to discriminative object parts inside an object proposal. During training, we fix the scaling factor to 0.5 and only accept scaling and translating in each spatial transformer. Thus, $A_{\theta}$ is given by
\begin{equation}
A_{\theta} = \\
\left[
	\begin{array}{ccc}
	0.5 & 0 & t_x \\
	0 & 0.5 & t_y \\
	\end{array}	
\right]
\end{equation}

where $\theta = [t_x, t_y]$ are the translation parameters that are predicted based on the localization network.

Taking the local context feature map $\bm{F}_{l\_fused} \in \mathbb{R}^{D \times S \times S}$ as input, each transformer in our object part attention module transforms and samples the input map to the output map $\bm{q} \in \mathbb{R}^{D \times S \times S}$. After normalization, the outputs of each transformer are concatenated with the local context feature to form a mid-level feature representation for an object proposal, defined as

\begin{equation}
	\bm{F}_{mid} = concat(\bm{F}_{l\_fused}, \ \bm{q}_i, \ ..., \ \bm{q}_N).
\end{equation}

where $\bm{q}_i$ is the output of the $i_{th}$ transformer and $N$ is the number of spatial transformers.

As shown in Figure \ref{fig:framework}, we use a $1 \times 1$ convolution layer after re-scaling to reduce the dimensions of $\bm{F}_{mid}$ from $S \times S \times (N \times D)$ to $S \times S \times D$, which is then fed to two fully connected layers to infer the final feature representation for the object proposal, denoted as $\bm{F}_L$.

\begin{figure*}
\centering
  \includegraphics[width=0.9\textwidth]{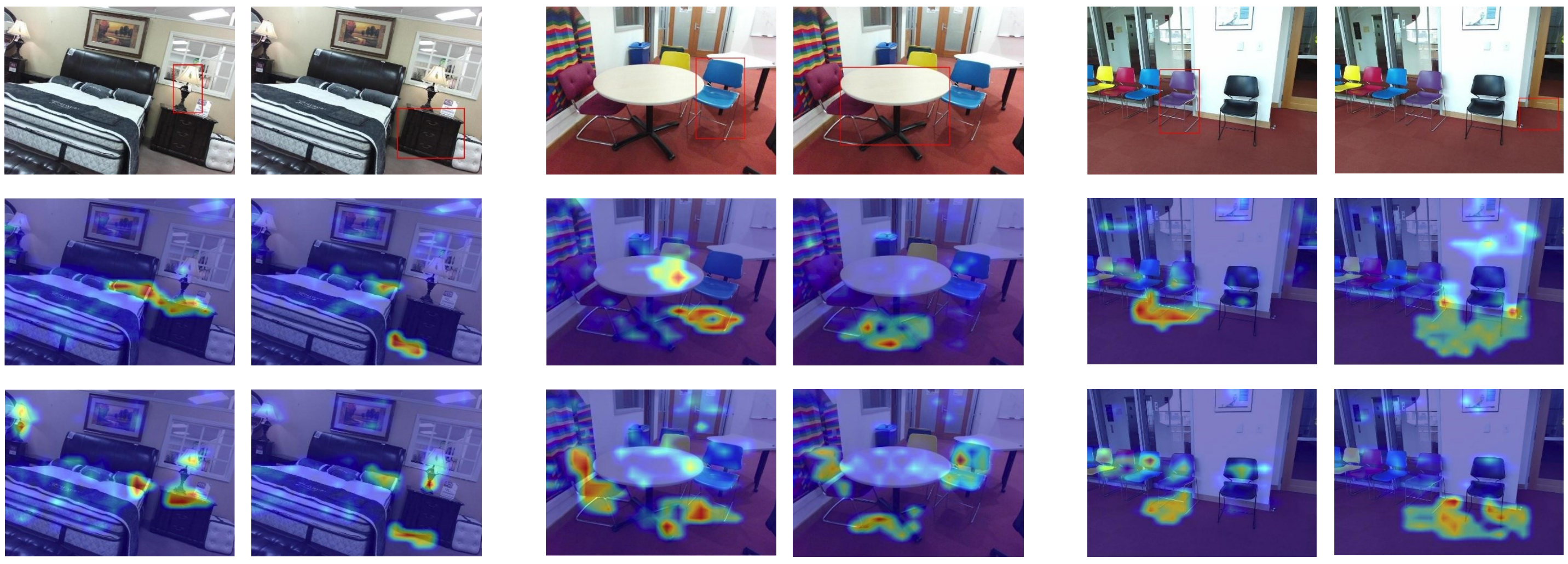}
  \caption{Illustration of the attentional weight maps generated by the attention-based global context modeling module. The top rows are the input images and region proposals. The middle and bottom rows are  the attentional weight maps generated by our model without context fusion and those with context fusion, respectively. The bottom two rows show that our model can perceive the most relevant regions to the given object proposal and that more useful regions can be acquired through context fusion. A detailed discussion can be found in section \ref{sec:visualization}.}
\label{fig:detection maps}
\end{figure*}

\subsection{Training Objective}
Denote $p = (p_0,...,p_L)$ as the predicted discrete probability distribution (per ROI) over $C+1$ categories and $t^*$ as the predicted bounding-box regression offsets. Given the obtained local and global context features ~$\bm{F}_L$ and~$\bm{F}_G$, $p$ and $t^*$ can be computed as follows:

\begin{equation}
	p = \textit{Softmax} \left( \textit{$f_{cls}$} \left( concat(\bm{F}_L, \bm{F}_G) \right) \right)
\end{equation}

\begin{equation}
	t^* = \textit{$f_{loc}$} \left( \bm{F}_L \right)
\end{equation}

where $Softmax(\cdot)$ indicates the softmax operation and $f_{cls}$ and $f_{loc}$ are two fully connected layers with $C+1$ units and $4 \times C$ units, respectively.

Note that we only incorporate local contextual information for bounding-box regression. Finally, we minimize an objective function following the multi-task loss given in Fast-RCNN \cite{fast-rcnn}, which is defined as

\begin{equation}
	L(p, u, t^u, v) = L_{cls}(p, u) + [u \geq 1] L_{loc}(t^u, v)
\end{equation}

where $u$ is the ground-truth label, $v$ is the regression target, $L_{cls}$ is the log loss for ground-truth class $u$, and $L_{loc}$ is the smooth $L_1$ loss proposed in \cite{fast-rcnn}. $[u \geq 1]$ evaluates to 1 when $u \geq 1$ and 0 otherwise. By convention, the background class is labeled as $u = 0$.

\begin{table*} \addtolength{\tabcolsep}{-2pt} \centering \renewcommand\arraystretch{1.5}
	\caption{\label{table:1} Detection results on SUNRGBD. AC-CNN* indicates our implementation of the RGB-D version of AC-CNN \cite{li2016attentive}. (w/o fusion) and (w/ fusion) denote without and with multi-modal context fusion, respectively.}
	\begin{tabular}{c|c|ccccccccccccccccccc}
	\toprule
	\textbf{\small{Method}}
	 	&\textbf{\small{mAP}}
	 	&\textbf{\tiny{bathtub}}
		&\textbf{\tiny{bed}}&\textbf{\tiny{bookshelf}}
		&\textbf{\tiny{box}}&\textbf{\tiny{chair}}
		&\textbf{\tiny{counter}}&\textbf{\tiny{desk}}
		&\textbf{\tiny{door}}&\textbf{\tiny{dresser}}
		&\textbf{\tiny{garbage\_bin}}&\textbf{\tiny{lamp}} 					&\textbf{\tiny{monitor}}&\textbf{\tiny{night\_stand}}
		&\textbf{\tiny{pillow}}&\textbf{\tiny{sink}}
		&\textbf{\tiny{sofa}}&\textbf{\tiny{table}}
		&\textbf{\tiny{tv}}&\textbf{\tiny{toilet}} \\
	\midrule
	
	\small{RGB-D RCNN \cite{gupta2014learning}}
		&\small{35.2}
		&\tiny{49.6}&\tiny{76.0}
		&\tiny{35.0}&\tiny{5.8}
		&\tiny{41.2}&\tiny{8.1}
		&\tiny{16.6}&\tiny{4.2}
		&\tiny{31.4}&\tiny{46.8}
		&\tiny{22.0}&\tiny{10.8}
		&\tiny{37.2}&\tiny{16.5}
		&\tiny{41.9}&\tiny{42.2}
		&\tiny{43.0}&\tiny{32.9}
		&\tiny{69.8} \\
		
	\small{ST(baseline) \cite{gupta2016cross}}
		&\small{43.8}
		&\tiny{65.3}&\tiny{83.0}
		&\tiny{54.4}&\tiny{14.4}
		&\tiny{46.9}&\tiny{14.6}
		&\tiny{23.9}&\tiny{15.3}
		&\tiny{41.3}&\tiny{51.0}
		&\tiny{32.1}&\tiny{36.8}
		&\tiny{46.6}&\tiny{23.4}
		&\tiny{43.9}&\tiny{61.3}
		&\tiny{48.7}&\tiny{50.5}
		&\tiny{79.4} \\
		
	\small{AC-CNN*}
		&\small{45.4}
		&\tiny{65.8}&\tiny{83.3}
		&\tiny{56.2}&\tiny{16.4}
		&\tiny{47.5}&\tiny{16.0}
		&\tiny{24.9}&\tiny{16.6}
		&\tiny{42.7}&\tiny{53.4}
		&\tiny{33.8}&\tiny{39.5}
		&\tiny{47.1}&\tiny{25.2}
		&\tiny{45.3}&\tiny{61.9}
		&\tiny{49.0}&\tiny{54.1}
		&\tiny{84.2} \\
	\midrule
	
	\small{Ours (w/o fusion)}
		&\small{46.9}
		&\tiny{68.2}&\tiny{85.7}
		&\tiny{56.0}&\tiny{17.3}
		&\tiny{49.8}&\tiny{17.1}
		&\tiny{25.2}&\tiny{16.9}
		&\tiny{43.5}&\tiny{54.2}
		&\tiny{35.5}&\tiny{40.7}
		&\tiny{49.4}&\tiny{26.1}
		&\tiny{\textbf{46.6}}&\tiny{66.3}
		&\tiny{52.0}&\tiny{56.5}
		&\tiny{84.3} \\
	\small{Ours (w/ fusion)}
		&\small{\textbf{47.5}}
		&\tiny{\textbf{69.0}}&\tiny{\textbf{86.1}}
		&\tiny{\textbf{57.9}}&\tiny{\textbf{18.2}}
		&\tiny{\textbf{50.3}}&\tiny{\textbf{17.4}}
		&\tiny{\textbf{26.8}}&\tiny{\textbf{17.3}}
		&\tiny{\textbf{44.4}}&\tiny{\textbf{54.4}}
		&\tiny{\textbf{35.6}}&\tiny{\textbf{40.5}}
		&\tiny{\textbf{49.8}}&\tiny{\textbf{26.7}}
		&\tiny{46.6}&\tiny{\textbf{67.2}}
		&\tiny{\textbf{52.9}}&\tiny{\textbf{56.7}}
		&\tiny{\textbf{84.9}} \\
	\bottomrule
	\end{tabular}
\end{table*}

\begin{table*} \addtolength{\tabcolsep}{-2pt} \centering  \renewcommand\arraystretch{1.5}
	\caption{\label{table:2} Detection results on NYUv2. AC-CNN* indicates our implementation of the RGB-D version of AC-CNN \cite{li2016attentive}. (w/o fusion) and (w/ fusion) denote without and with multi-modal context fusion, respectively.}
	\begin{tabular}{c|c|ccccccccccccccccccc}
	\toprule
	\textbf{\small{Method}}
	 	&\textbf{\small{mAP}}
	 	&\textbf{\tiny{bathtub}}
		&\textbf{\tiny{bed}}&\textbf{\tiny{bookshelf}}
		&\textbf{\tiny{box}}&\textbf{\tiny{chair}}
		&\textbf{\tiny{counter}}&\textbf{\tiny{desk}}
		&\textbf{\tiny{door}}&\textbf{\tiny{dresser}}
		&\textbf{\tiny{garbage\_bin}}&\textbf{\tiny{lamp}} 					&\textbf{\tiny{monitor}}&\textbf{\tiny{night\_stand}}
		&\textbf{\tiny{pillow}}&\textbf{\tiny{sink}}
		&\textbf{\tiny{sofa}}&\textbf{\tiny{table}}
		&\textbf{\tiny{tv}}&\textbf{\tiny{toilet}} \\
	\midrule
	
	\small{RGB-D RCNN \cite{gupta2014learning}}
		&\small{32.5}
		&\tiny{22.9}&\tiny{66.5}
		&\tiny{21.8}&\tiny{3.0}
		&\tiny{40.8}&\tiny{37.6}
		&\tiny{10.2}&\tiny{20.5}
		&\tiny{26.2}&\tiny{37.6}
		&\tiny{29.3}&\tiny{43.4}
		&\tiny{39.5}&\tiny{37.4}
		&\tiny{24.2}&\tiny{42.8}
		&\tiny{24.3}&\tiny{37.2}
		&\tiny{53.0} \\
		
	\small{ST(baseline) \cite{gupta2016cross}}
		&\small{49.1}
		&\tiny{50.6}&\tiny{81.0}
		&\tiny{52.6}&\tiny{5.4}
		&\tiny{53.0}&\tiny{56.1}
		&\tiny{21.0}&\tiny{34.6}
		&\tiny{57.9}&\tiny{46.2}
		&\tiny{42.5}&\tiny{62.9}
		&\tiny{54.7}&\tiny{49.1}
		&\tiny{50.0}&\tiny{65.9}
		&\tiny{31.9}&\tiny{50.1}
		&\tiny{68.0} \\
		
	\small{AC-CNN*}
		&\small{50.2}
		&\tiny{52.2}&\tiny{82.4}
		&\tiny{52.5}&\tiny{8.6}
		&\tiny{54.8}&\tiny{57.3}
		&\tiny{22.7}&\tiny{34.1}
		&\tiny{58.1}&\tiny{46.5}
		&\tiny{42.9}&\tiny{63.6}
		&\tiny{55.2}&\tiny{49.7}
		&\tiny{51.4}&\tiny{66.8}
		&\tiny{33.5}&\tiny{51.8}
		&\tiny{70.4} \\
	
	\midrule	
	\small{Ours (w/o fusion)}
		&\small{51.9}
		&\tiny{55.2}&\tiny{83.4}
		&\tiny{\textbf{54.2}}&\tiny{9.4}
		&\tiny{55.1}&\tiny{58.5}
		&\tiny{24.0}&\tiny{35.9}
		&\tiny{58.3}&\tiny{46.6}
		&\tiny{44.8}&\tiny{65.7}
		&\tiny{57.0}&\tiny{\textbf{52.7}}
		&\tiny{53.6}&\tiny{68.4}
		&\tiny{\textbf{35.3}}&\tiny{54.8}
		&\tiny{73.9} \\
		
	\small{Ours (w/ fusion)}
		&\small{\textbf{52.3}}
		&\tiny{\textbf{55.6}}&\tiny{\textbf{83.9}}
		&\tiny{54.0}&\tiny{\textbf{9.8}}
		&\tiny{\textbf{55.4}}&\tiny{\textbf{59.2}}
		&\tiny{\textbf{24.1}}&\tiny{\textbf{36.3}}
		&\tiny{\textbf{58.5}}&\tiny{\textbf{47.2}}
		&\tiny{\textbf{45.0}}&\tiny{\textbf{65.8}}
		&\tiny{\textbf{57.6}}&\tiny{52.7}
		&\tiny{\textbf{53.8}}&\tiny{\textbf{69.1}}
		&\tiny{35.0}&\tiny{\textbf{56.9}}
		&\tiny{\textbf{74.7}} \\
	\bottomrule
	\end{tabular}
\end{table*}

\section{Experimental Results}

\subsection{Experimental Settings}

\textbf{Datasets and Evaluation Metrics:} We evaluate our model on two RGB-D datasets: SUNRGBD \cite{song2015sun} and NYUv2 \cite{silberman2012indoor}. The SUNRGBD and NYUv2 datasets contain 10335 and 1449 RGB-D images, respectively, and are divided into \textit{train} and \textit{test} subsets. We adopt \textit{Average Precision} (AP) and \textit{mean of Average Precision} (mAP) following the PASCAL challenge protocols as our evaluation metrics.

\textbf{Implementation Details:} In our experiments, we implement our model based on Fast R-CNN~\cite{fast-rcnn}, an open-source framework for traditional RGB object detection built on the Caffe platform~\cite{jia2014caffe}. We utilize the network architecture from Gupta\textit{et al}~\cite{gupta2016cross} as our basic CNN network structure for convolutional feature map extraction. All the newly added fully connected and convolutional layers are randomly initialized with a zero-mean Gaussian distribution with standard deviations of 0.01 and 0.001. The recurrent attention model consists of 4 stacked LSTM units with shared parameters. All the parameters of the LSTM units are initialized based on the xavier algorithm~\cite{glorot2010understanding}.

\begin{table} \centering \renewcommand\arraystretch{1.2}
	\caption{\label{table:4} Comparison of different LSTM settings utilized in the attention-based global context sub-module. The experiments are conducted on SUNRGBD. (2 $\times$ LSTM) denotes that there are 2 stacked LSTM units in the global contextualized sub-network.}
	\begin{tabular}{c|c}
	\toprule
	\textbf{LSTM Settings}&\textbf{mAP} \\
	\midrule
	Ours (2 $\times$ LSTM)  & 45.4 \\
	Ours (3 $\times$ LSTM)  & 46.0 \\	
	Ours (4 $\times$ LSTM)  & \textbf{46.2} \\
	Ours (5 $\times$ LSTM)  & \textbf{46.2} \\
	\bottomrule
	\end{tabular}
\end{table}

We apply Stochastic Gradient Decent (SGD) to fine tune our model. Each SGD mini-batch is composed of 128 randomly sampled object proposals from 2 randomly chosen images. In each mini-batch, we select 25\% of the ROIs as foreground from object proposals that have intersection over union (IoU) overlap with a ground-truth bounding box of at least 0.5. The remaining ROIs are sampled from object proposals that have a maximum IoU with ground truth in the interval $[0.1, 0.5)$ and act as background with ground truth label $u = 0$. During training, images are horizontally flipped with a probability of 0.5 for data augmentation, and no other augmentation is used. We run SGD for approximately 10 epochs on the training set to fine tune the network parameters. The momentum is set to 0.9, and the learning rate is initialized to 0.001 and decreased by 10 every 4 epochs. It takes approximately 1.5 days to train our model on a single NVIDIA GeForce GTX TITAN X GPU with 12 GB of memory. 

It costs approximately 10 GB of GPU memory to train our model. The average training time for each iteration is approximately 1.23 seconds. However, the testing process is particularly efficient and takes approximately 0.58 seconds (excluding object proposal extraction) to process one image.

\subsection{Performance Comparisons}
\textbf{RGB-D Datasets}: We compare our proposed method against recent state-of-the-art RGB-D object detection methods, including rich image and depth feature-based RGB-D object detection~\cite{gupta2014learning} and the supervision-transfer-based model~\cite{gupta2016cross}. Moreover, to better validate the superiority of the attention-based global context and fine-grained object part attention on RGB-D datasets, we also implement an RGB-D version (denoted as AC-CNN*) of the AC-CNN model proposed in~\cite{li2016attentive} for comparison. AC-CNN follows a similar idea to our proposed method but incorporates fixed global and local attentive contexts to assist in improving the object detection performance. In the implementation, we apply the Fast RCNN \cite{fast-rcnn} framework based on AlexNet \cite{alexnet} to the depth modality for proposal classification and bounding-box position regression. The final results are obtained by averaging the results from the RGB modality and depth modality. For fair comparison, we also apply the same depth modality processing as in AC-CNN* to our model; we call this custom model RGB-D detection without cross-modal fusion~(denoted as w/o fusion).

Table \ref{table:1} and Table \ref{table:2} illustrate the object detection results of our model, AC-CNN*, and the other two state-of-the-art RGB-D object detection models on the SUNRGBD and NYUv2 datasets. As shown in the table, our proposed method obtains state-of-the-art mAP scores of 47.5\% and 52.3\% on SUNRGBD and NYUv2, which outperforms the ST model \cite{gupta2016cross} by 3.7\% and 3.2\%, respectively. The improvements validate the effectiveness of our model in RGB-D object detection by incorporating the proposed attention-based global context and fine-grained attentional object parts learned from the fused cross-modal context.  Furthermore, our model~(Ours (w/o fusion)) gains 1.5\% and 1.7\% improvements in mAP scores over AC-CNN* on the  SUNRGBD and NYUv2 datasets,  respectively, and achieve better detection results on most of the categories.

\textbf{RGB Dataset}: To compare our model with the AC-CNN model~\cite{li2016attentive} in a more equitable way, we remove the depth modality from our model and perform an extra evaluation on PASCAL VOC 2007, which contains 9963 RGB images. Specifically, we implement a variant of our model (denoted as Ours*) that performs global context modeling and object part attention only on the RGB modality without incorporating information from the depth modality. As shown in Table~\ref{table:3}. Our model outperforms the baseline FRCN \cite{fast-rcnn} and AC-CNN \cite{li2016attentive} by 3.6\% and 1.2\% in terms of mAP scores, respectively. The improvement on the RGB dataset as well as the favorable results achieved for RGB-D object detection well demonstrate the superiority of the proposed attention-based global context and fine-grained object part attention over the fixed global context and multi-scale local context proposed in \cite{li2016attentive}. Table \ref{table:voc_2012} provides the comparisons of the proposed method with several state-of-the-art methods~\cite{ravishankar2008multi, liu2016ssd, shen2017dsod, bell2016inside, dai2016r} on PASCAL VOC 2012. It can be observed that our model obtains an mAP score of 76.7\%, which outperforms the baseline model by 2.9\%. Our model also achieves competitive results compared with the state-of-the-art methods, which validates the effectiveness of the proposed method.

\begin{table} \centering \renewcommand\arraystretch{1.5}
	\caption{\label{table:5} Comparison of different STN settings utilized in fine-grained object part attention sub-module. The experiments are conducted on SUNRGBD. (2 $\times$ STN) indicates that there are 2 parallel spatial transformers in the local contextualize sub-network.}
	\begin{tabular}{c|c}
	\toprule
	\textbf{STN Settings}&\textbf{mAP} \\
	\midrule
	Ours (1 $\times$ STN) & 45.7 \\
	Ours (2 $\times$ STN) & \textbf{46.3} \\
	Ours (3 $\times$ STN) & 46.0 \\
	\bottomrule
	\end{tabular}
\end{table}

\subsection{Ablation Studies}
\label{sec:ablation}
In this subsection, we show the effectiveness and necessity of each component in our proposed model and also demonstrate the effectiveness of the network design. 

\begin{figure*}
\centering
  \includegraphics[width=0.9\textwidth]{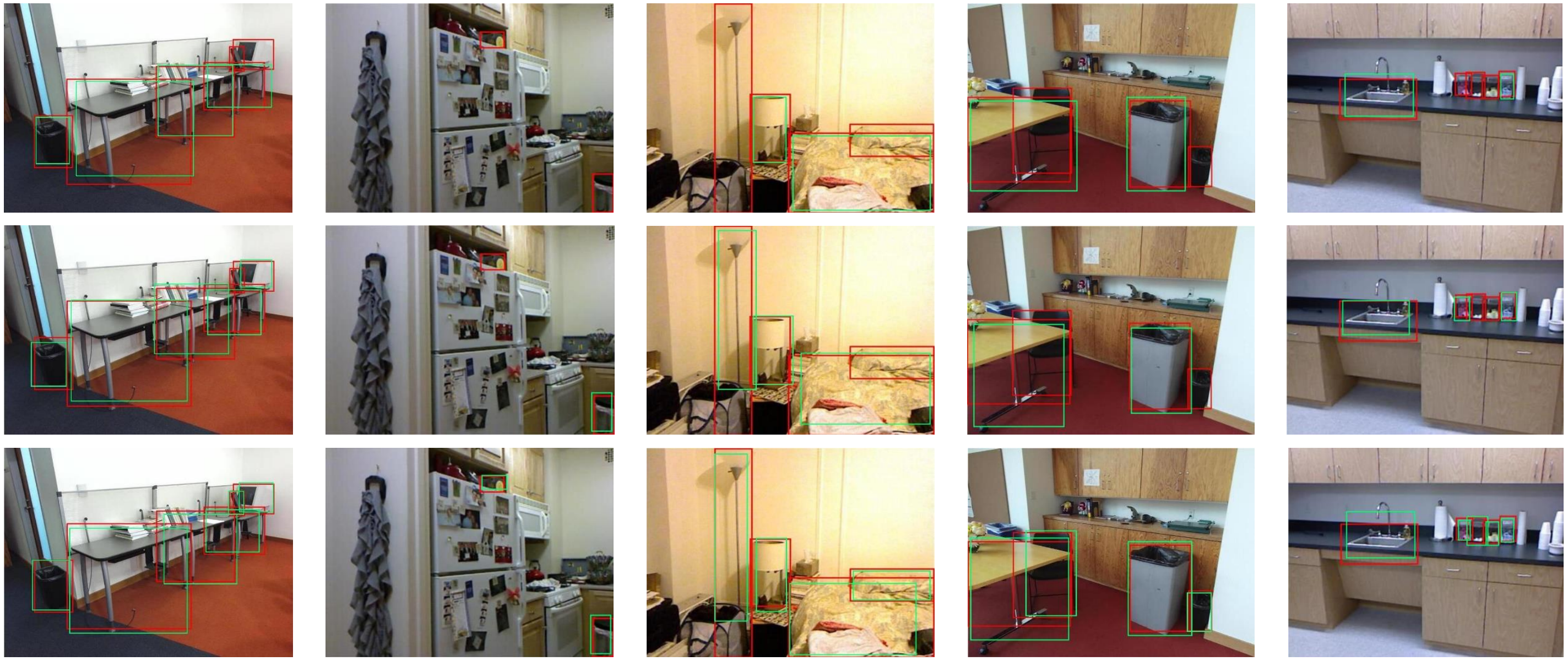}
	\caption{Comparison of detection results produced by ST
		\cite{gupta2016cross} (top row), AC-CNN 
		\cite{li2016attentive} (middle row) and our model (bottom row). 	
		The red and green rectangles indicate the ground-truth bounding box and the predicted results, respectively.}
\label{fig:detection results}
\end{figure*}

\begin{table*} \addtolength{\tabcolsep}{-3pt} \renewcommand\arraystretch{1.5}
	\caption{\label{table:3} Detection results on VOC 2007. Ours* denotes a variant of our model in which we incorporate only RGB information for object detection}
	\begin{tabular}{c|c|cccccccccccccccccccc}
	\toprule
	\textbf{\small{Method}}
	 	&\textbf{\small{mAP}}
	 	&\textbf{\scriptsize{aero}}&\textbf{\scriptsize{bike}}
	 	&\textbf{\scriptsize{bird}}&\textbf{\scriptsize{boat}}
		&\textbf{\scriptsize{bottle}}&\textbf{\scriptsize{bus}}
		&\textbf{\scriptsize{car}}&\textbf{\scriptsize{cat}}
		&\textbf{\scriptsize{chair}}&\textbf{\scriptsize{cow}}
		&\textbf{\scriptsize{table}}&\textbf{\scriptsize{dog}}
		&\textbf{\scriptsize{horse}}&\textbf{\scriptsize{mbike}}
		&\textbf{\scriptsize{person}}&\textbf{\scriptsize{plant}}
		&\textbf{\scriptsize{sheep}}&\textbf{\scriptsize{sofa}}
		&\textbf{\scriptsize{train}}&\textbf{\scriptsize{tv}} \\
	\midrule
	
	\small{FRCN(Baseline) \cite{fast-rcnn}}
		&\small{70.0}
		&\scriptsize{77.0}&\scriptsize{78.1}
		&\scriptsize{69.3}&\scriptsize{59.4}
		&\scriptsize{38.3}&\scriptsize{81.6}
		&\scriptsize{78.6}&\scriptsize{86.7}
		&\scriptsize{42.8}&\scriptsize{78.8}
		&\scriptsize{68.9}&\scriptsize{84.7}
		&\scriptsize{82.0}&\scriptsize{76.6}
		&\scriptsize{69.9}&\scriptsize{31.8}
		&\scriptsize{70.1}&\scriptsize{74.8}
		&\scriptsize{80.4}&\scriptsize{70.4} \\
	
	\small{AC-CNN \cite{li2016attentive}}
		&\small{72.4}
		&\scriptsize{79.1}&\scriptsize{79.2}
		&\scriptsize{71.9}&\scriptsize{\textbf{61.0}}
		&\scriptsize{43.2}&\scriptsize{83.0}
		&\scriptsize{81.4}&\scriptsize{87.7}
		&\scriptsize{50.0}&\scriptsize{82.1}
		&\scriptsize{73.6}&\scriptsize{83.4}
		&\scriptsize{84.2}&\scriptsize{77.5}
		&\scriptsize{72.0}&\scriptsize{35.8}
		&\scriptsize{71.9}&\scriptsize{74.7}
		&\scriptsize{\textbf{85.8}}&\scriptsize{71.0} \\
	
	\small{Ours*}
		&\small{\textbf{73.6}}
		&\scriptsize{\textbf{81.0}}&\scriptsize{\textbf{80.2}}
		&\scriptsize{\textbf{72.4}}&\scriptsize{60.5}
		&\scriptsize{\textbf{45.3}}&\scriptsize{\textbf{84.1}}
		&\scriptsize{\textbf{82.8}}&\scriptsize{\textbf{88.0}}
		&\scriptsize{\textbf{51.6}}&\scriptsize{\textbf{82.5}}
		&\scriptsize{\textbf{74.8}}&\scriptsize{\textbf{85.7}}
		&\scriptsize{\textbf{84.9}}&\scriptsize{\textbf{79.6}}
		&\scriptsize{\textbf{72.2}}&\scriptsize{\textbf{36.9}}
		&\scriptsize{\textbf{72.1}}&\scriptsize{\textbf{76.8}}
		&\scriptsize{85.5}&\scriptsize{\textbf{74.3}} \\
			
	\bottomrule
	\end{tabular}
\end{table*}

\textbf{Contribution of Each Component in CMAC model}: As described in Section~\ref{sec:framework}, our proposed CMAC model
consists of three newly added sub-networks on the top of deep feature representation, including cross-modal feature fusion, attention-based global context modeling and fine-grained  object part attention, which are employed to incorporate the strong correlation between different modalities and capture the global and local contextual information, respectively. We investigate the contributions of each component by gradually applying each sub-network to the object detection.  Table~\ref{table:6} shows that 2.5\% and 1.8\% improvements in mAP scores over the baseline model are obtained using only fine-grained object part attention. Similar improvements of 2.4\% and 2.2\% on SUNRGBD and NYUv2 can be observed when only incorporating attention-based global context modeling. The better performance achieved by exploiting both global context features and discriminative object parts evidences the complementarity of the two sub-networks. Furthermore, incorporating cross-modal feature fusion into our detection framework brings an extra performance increase of 0.6\% and 0.4\% on SUNRGBD and NYUv2, respectively. The above experimental results and analysis well demonstrate the effectiveness of  each component in our proposed CMAC framework.

\textbf{Comparison of Diverse Global Context Modeling}: To validate the effectiveness of our attention-based global context, which is generated based on a recurrent model, we compare our model with two variants: the global average pooling method in which the global contextual information is produced by applying the average pooling operation to the extracted feature map, and AC-CNN, which utilizes an attention-based recurrent model to generate the fixed global context. We conduct experiments on the SUNRGBD dataset, and the results are listed in Table~\ref{table:8}. No local context is used during these experiments. It can be observed that our model outperforms the global averaging pooling method and AC-CNN by 1.9\% and 1.5\%, respectively. Simply averaging the features of all regions may introduce both background and inter-class noise, which may deteriorate the object detection performance. Although background noise can be overcome by AC-CNN, which generates a fixed attention map for global context feature extraction and benefits the proposal classification, AC-CNN still suffers from a decreased performance caused by inter-class noise~(e.g., regions that are beneficial for desk classification might provide noisy information to garbage\_bin classification). Note that our attention map for global context weighting is generated according to the diverse contents of each  ROI feature and can be optimized to attend to the most effective regions related to the input content. The results shown in Table~\ref{table:8} verify that our model performs better in mitigating both background and inter-class noise by incorporating global context and thus greatly enhances the accuracy of object detection.

\textbf{Effectiveness of LSTM Settings}: In our proposed CMAC model, we have employed a recurrent model to exploit the attentional global context, in which multiple stacked LSTM units are utilized to generate the attentional weight map in an iterative manner. To investigate the effectiveness of different LSTM settings, we implement several variants, whereby the recurrent model is constructed with different numbers (2 to 5) of LSTM units. The experimental results are listed in Table~\ref{table:4}. As shown in the table, the mAP metric increases by 0.6\% and 0.8\% when the number of stacked LSTM units is increased from $2$ to $3$ and $4$, respectively. When this number reaches or exceeds $5$, no significant performance boosts are achieved, indicating that our model can obtain better context information through recurrent iterations and will converge quickly. We believe that good performance can be obtained in complicated images through more recurrent iterations. 

\textbf{Effectiveness of STN Settings}: In the proposed method, we adopt several parallel multiple transform networks (STNs) to attend to discriminative object parts inside an object proposal. To investigate the most effective STN setting, we implement several variants whereby the fine-grained object parts are inferred from different numbers (2 to 4) of spatial transformers. As shown in Table~\ref{table:5}, the detection performance increases from 43.8\% (baseline) to 45.7\% and 46.3\% with $1$ and $2$ spatial transformers, respectively, which indicates that STNs are able to mine discriminative object parts to enhance the local feature representation. However, increasing the number of spatial transformers does not always bring about a better performance. We observe a 3\% decrease in mAP when increasing the number of spatial transformers from 2 to 3, indicating that the STNs may start to enroll confusing object parts after most of the discriminative parts have been detected.

\begin{table}[h] \centering
	\caption{\label{table:voc_2012}  PASCAL VOC 2012 test detection results. \textbf{07+12+S:} 07 trainval + 12 trainval + segmentation labels, \textbf{07++12:} 07
trainval + 07 test + 12 trainval}
	\begin{tabular}{c|c|c}
	\toprule
	\textbf{Method}&\textbf{data}&\textbf{mAP} \\
	\midrule
	Faster-Rcnn~\cite{ren2015faster} (Baseline) & 07++12  & 73.8 \\
	Multi-stage~\cite{li2017multi}     & 07++12  & 74.9 \\
	SSD300~\cite{liu2016ssd}                    & 07++12  & 75.8 \\
	DOSD~\cite{shen2017dsod}                    & 07++12  & 76.3 \\
	ION~\cite{bell2016inside}                   & 07+12+S & 76.4 \\
	R-FCN multi-scale~\cite{dai2016r}           & 07++12  & \textbf{77.6} \\  
	\midrule
	ours                   & 07++12  & 76.7 \\
	\bottomrule
	\end{tabular}
\end{table}

\subsection{Visualization} 
\label{sec:visualization}
In this subsection, we present some visual comparisons of the RGB-D object detection results as well as some visual effects of the attentional weight maps generated by our global context modeling component. Figure~\ref{fig:detection results} shows some detection results of the ST~\cite{gupta2016cross} model, the AC-CNN \cite{li2016attentive} model and our model. It can be observed that our model performs best in detecting small and occluded objects~(e.g., monitor, box, garbage\_bin and the occluded chair). Furthermore, as shown in the third column, our proposed method is also more robust to appearance-similar instances because of the fusion of the geometry context~(e.g., the pillow with similar texture to the bed). Figure~\ref{fig:detection maps} demonstrates the attentional weight maps generated by our model without~(middle row) and with~(bottom row) context fusion. Obviously, our attentional model is able to perceive regions most relevant to the specific object proposal,~\textit{i.e.,} a lamp is likely to be placed on top of a night stand near a bed, and a night stand is also likely to be placed on the floor near a bed and often co-occurs with a lamp. Moreover, our model obtains more accurate attentional weight maps by fusing information from both RGB and depth modalities since the depth image can provide geometric information. For example, our model is capable of attending to the chairs near the target chair, as they share similar geometric structures. The last column in Fig.~\ref{fig:detection maps} shows that our model will attend to the background regions when the proposal does not contain objects, which helps in making correct classifications.

\section{Conclusion}
In this paper, we have introduced an approach to effectively learn the cross-modal attentive context for RGBD object detection. In our model, the contextual representations from different sources (\textit{i.e.,} RGB and depth modalities) are fused in the cross-modal feature fusion module. Based on the fused local and global feature, a recurrent attention model including several stacked LSTM units is employed to capture a global context that is closely related to the object proposal. Furthermore, our model adopts several parallel spatial transformers, which learn to attend to discriminative parts inside each object proposal, to generate the enhanced local context information. Extensive experiments and state-of-the-art detection results on SUNRGBD and NYUv2 well demonstrate the effectiveness of our model in exploiting contextual information.


%

%
%
%
%
%


\ifCLASSOPTIONcaptionsoff
  \newpage
\fi




{\small
\bibliographystyle{IEEEtran}
\bibliography{bare_jrnl}
}

\end{document}